\documentclass[journal, 10pt]{IEEEtran}
\pdfoutput=1
\IEEEoverridecommandlockouts
\usepackage{cite}
\usepackage{amsmath,amssymb,amsfonts}
\usepackage{algorithmic}
\usepackage{graphicx}
\usepackage{textcomp}
\usepackage{xcolor}
\usepackage[ruled,vlined]{algorithm2e}
\usepackage{multirow}
\usepackage{xspace}
\usepackage{subfigure}
\usepackage{graphicx} 
\usepackage{multirow}
\usepackage{url}
 \usepackage{enumitem}
 \usepackage{diagbox}
 \usepackage[skip=0.5\baselineskip,font=footnotesize]{caption}

\ifCLASSINFOpdf
\else
\fi



\begin{document}
%
\title{EvoLP: Self-Evolving Latency Predictor for Model Compression in Real-Time Edge Systems}
%
%
%

\author{
Shuo~Huai,~Hao~Kong,~Shiqing~Li,~Xiangzhong~Luo, Ravi~Subramaniam,~Christian~Makaya,~Qian~Lin~and~Weichen~Liu
\thanks{\copyright~2024 IEEE. Personal use of this material is permitted. Permission from IEEE must be obtained for all other uses, in any current or future media, including reprinting/republishing this material for advertising or promotional purposes, creating new collective works, for resale or redistribution to servers or lists, or reuse of any copyrighted component of this work in other works. This is the author's accepted version of the article published in IEEE Embedded Systems Letters, vol. 16, no. 2, pp. 174-177, June 2024, DOI: 10.1109/LES.2023.3321599.}
\thanks{

S. Huai, H. Kong, S. Li, X. Luo, and W. Liu (Corresponding author, liu@ntu.edu.sg) are with the School of Computer Science and Engineering, Nanyang Technological University, Singapore; S. Huai and H. Kong are also with the HP-NTU Digital Manufacturing Corporate Lab, Nanyang Technological University, Singapore; R. Subramaniam, C. Makaya, and Q. Lin are with HP Inc., Palo Alto, California, USA.

This work is partially supported under the RIE2020 Industry Alignment Fund – Industry Collaboration Projects (IAF-ICP) Funding Initiative, as well as cash and in-kind contribution from the industry partner, HP Inc., through the HP-NTU Digital Manufacturing Corporate Lab (I1801E0028), and partially supported by the Ministry of Education, Singapore, under its Academic Research Fund Tier 2 (MOE2019-T2-1-071), and Nanyang Technological University, Singapore, under its NAP }


\vspace{-1.23cm}
}

%
%

\markboth{IEEE EMBEDDED SYSTEMS LETTERS}
{Shell \MakeLowercase{\textit{et al.}}: Bare Demo of IEEEtran.cls for IEEE Journals}
%



\maketitle

\begin{abstract}
Edge devices are increasingly utilized for deploying deep learning applications on embedded systems.
The real-time nature of many applications and the limited resources of edge devices necessitate latency-targeted neural network compression. However, measuring latency on real devices is challenging and expensive. Therefore, this letter presents a novel and efficient framework, named EvoLP, to accurately predict the inference latency of models on edge devices. This predictor can evolve to achieve higher latency prediction precision during the network compression process. Experimental results demonstrate that EvoLP outperforms previous state-of-the-art approaches by being evaluated on three edge devices and four model variants. Moreover, when incorporated into a model compression framework, it effectively guides the compression process for higher model accuracy while satisfying strict latency constraints. We open source EvoLP at \textcolor{blue}{\url{https://github.com/ntuliuteam/EvoLP}}.

\end{abstract}


%

%
%
%
%

 
\section{Introduction}
\label{section:introduction}

\IEEEPARstart{D}{eep} Neural Networks (DNNs) have been widely deployed on edge devices to eliminate the issues of data confidentiality breaches and unstable network bandwidth in accessing cloud servers \cite{DLSurvey}. However, current models grow in complexity to improve accuracy, rendering them unsuitable for resource-limited edge devices. Neural architecture search (NAS) can design customized DNN models for edge devices, but it is time-consuming and requires heavy engineering efforts \cite{luo2022surgenas}. Model compression is promising to improve DNNs' efficiency on edge devices. Many DNN applications, such as virtual/augmented reality, and autonomous driving, demand strict response latency, thus, the inference latency of DNNs should be treated as a hard constraint in model compression.

Model compression based on real latency provides additional benefits from better exploration of hardware features \cite{yang2018netadapt}. However, traditionally, considering latency as a compression objective requires interaction with the physical device to obtain the real latency, which is time-consuming. Each latency measurement process may take several minutes \cite{ChamNet}, making it prohibitively expensive in large design spaces. Some approaches are proposed to avoid the high cost of measuring latency, including using look-up tables (LUTs) \cite{ChamNet,yang2018netadapt} or {\color{black}hardware simulators \cite{gholami2018squeezenext, li2021flash} to predict the latency. However, LUTs cannot predict the latency of new models that are not recorded in the table; and it is difficult to build hardware simulators by analyzing the device resource and scheduling algorithms due to the opacity of most existing hardware.
Thus, LUTs and  hardware simulators are unsuitable for various existing edge devices and model compression.}


Recently, some learning/regression-based methods \cite{nnmeter, li2021towards} are proposed to estimate the inference latency of models on different devices. They first sample some model layers and measure them on the device to obtain precise latency, creating a dataset. Then, these methods use machine learning or regression analysis techniques to derive the patterns from the dataset to predict the latency of unseen models. However, the sampling and measuring remain expensive and can take up to 4.4 days for nn-Meter \cite{nnmeter}.  Furthermore,  unlike NAS,  during model compression, the number of input channels and output channels of each layer changes continuously, thus the search space is much huger, rendering these methods unaffordable. 


To address the aforementioned problems, we present a novel framework to accurately predict the model inference latency upon targeted hardware utilizing Multi-Layer Perceptron (MLP) neural network \cite{riedmiller2014multi} and self-evolution scheme. Specifically, the main contributions of this letter are summarized:


\begin{enumerate}[label={\arabic*)}, noitemsep]
\item We present a novel and efficient latency predictor, named EvoLP, to accurately estimate the inference latency of DNN models on diverse edge devices. 

\item We reduce the search space to include only the to-be-compressed model and the proposed method can automatically analyze this model to generate a small number of samples, greatly improving the building efficiency. 

\item We proposed an evolution scheme for our predictor during model compression, allowing it to evolve for higher precision without interrupting the compression process. 

\item We evaluate EvoLP on three edge devices and show that it achieves much higher prediction accuracy compared to existing approaches. And we also demonstrate the benefits of EvoLP in the model compression technique.
\end{enumerate}
\section{Background \& Related Works}
%

Complex models require compression to enable efficient inference on low-power and resource-limited edge devices \cite{DLSurvey}. 
Recent researches \cite{huai2023latency} propose to compress the model with one training process to reduce the training overhead. When used in real-time edge systems, it is essential to ensure that the compressed model meets the system latency constraint.
Meanwhile, as introduced in Section \ref{section:introduction}, the costly on-device latency measurement must be replaced with latency predictors.

Latency predictors based on LUTs \cite{ChamNet} and hardware simulators \cite{gholami2018squeezenext} are unsuitable in compressing models for various real-time edge systems. Also, predictors based on floating point operations (FLOPs) tend to be highly inaccurate as they do not account for runtime optimization of various operations \cite{nnmeter}. Meanwhile, recent predictors, such as  nn-Meter \cite{nnmeter} and TALM \cite{li2021towards}, can predict kernel-level latency while considering some optimizations of different operations during runtime. However, these methods require lots of sampling for training the predictor, especially used for model compression.

DNN models are typically composed of a limited set of operations, such as convolution, batch normalization, activation functions, multiplication, etc. However, the parameter configuration range of each operation is infinite \cite{paszke2019pytorch}. Therefore, even if only predicting the latency of each operation, it poses a significant challenge due to the extensive search space involved. Specifically, considering the commonly used configuration range of convolution in all layers of different models, which encompasses input size ($1-299$), input channel ($1-1024$), kernel size ($1-7$), number of kernels ($1-1024$), stride ($1-3$), and padding ($0-1$), the search space is approximately $1.3 \times 10^{10}$. Consequently, creating a training set from such a massive search space is highly challenging, and the prediction performance is typically poor when dealing with unseen configurations. Thus, in this letter, we propose to generate a small number of samples for measuring by only considering the model under compression, which accelerates the building process. Moreover, the proposed predictor can evolve during the compression process to achieve higher precision, particularly near the latency constraint.

\section{Self-Evolving Latency Predictor}
Traditional latency predictors estimate the latency of models by sampling from the total search space and learning their patterns, but they present two major drawbacks, especially used for model compression. Firstly, the search space is extensive, and numerous samples must be collected, leading to the time cost of measuring these samples on the device being prohibitively expensive. Also, the large search space poses a significant challenge in learning the latency patterns and decreases the precision. Secondly, while some existing methods consider model runtime optimization, some runtime features remain unaccounted for. For instance, the time of data transfer between layers, the input and output time, the parallel processing of branches, etc. In light of these two drawbacks, this letter proposes a model-specific self-evolving latency predictor to efficiently and accurately predict the on-device latency of a model during model compression. 

\begin{figure}[!t]
    \centering
    \captionsetup{format=plain,labelfont={color=black}}
    \includegraphics[trim={0pt 11.5pt 0pt 0pt},clip, width=0.48\textwidth]{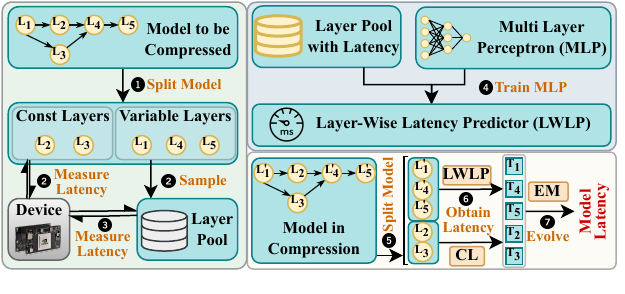}
    \caption{\textcolor{black}{Overview of our self-evolving latency predictor, including model sampling (left), latency predictor generation (right-top), and latency predictor evolution (right-bottom). The same numbers within circles denote processes that can be parallel. CL: Latency of const layers; EM: Evolution matrix. }}
    \label{fig:overview}
         \vspace{-0.5cm}
\end{figure}

Our latency predictor consists of three processes: {\color{black}model sampling (left), latency predictor generation (right-top), and latency predictor evolution (right-bottom)}, as shown in Fig.~\ref{fig:overview}. The first two processes are for constructing the predictor, {\color{black}and they reduce the search space to the model under compression}, improving both the building efficiency and prediction precision of the latency predictor. The last process is for the evolution scheme and can efficiently learn various model runtime features on the device, enabling accurate prediction of models with varying degrees of compression. We provide details of each phase in the following sections and demonstrate the usage of our latency predictor in model compression.

The first phase is the model sampling and we only need to know the architecture of the model to be compressed without training. 
{\color{black} Then according to the latency variation throughout the compression process, the layers are categorized into variable layers and const layers. Specifically, if the latency variation of a layer during the compression procedure is trivial, that layer is identified as a const layer. Conversely, it is classified as a variable layer. 
On-device latency is directly measured and recorded for const layers. For variable layers, we will sample according to the configuration ranges of each layer.}
%
As models only include limited layers, the sampling is highly efficient, taking only a few milliseconds. 
After sampling, we construct a layer pool that comprises thousands of single-layer models with varying operations (e.g., Convolution, Pooling) and configuration sizes, to train the latency predictor. The latency of a single layer typically falls within the range of a few milliseconds, and therefore, the process of measuring all models in the pool is completed within a few hours.

\begin{figure*}[!t]
    \centering

    \includegraphics[trim={0pt 6pt 0pt 0pt},clip, width=0.94\textwidth]{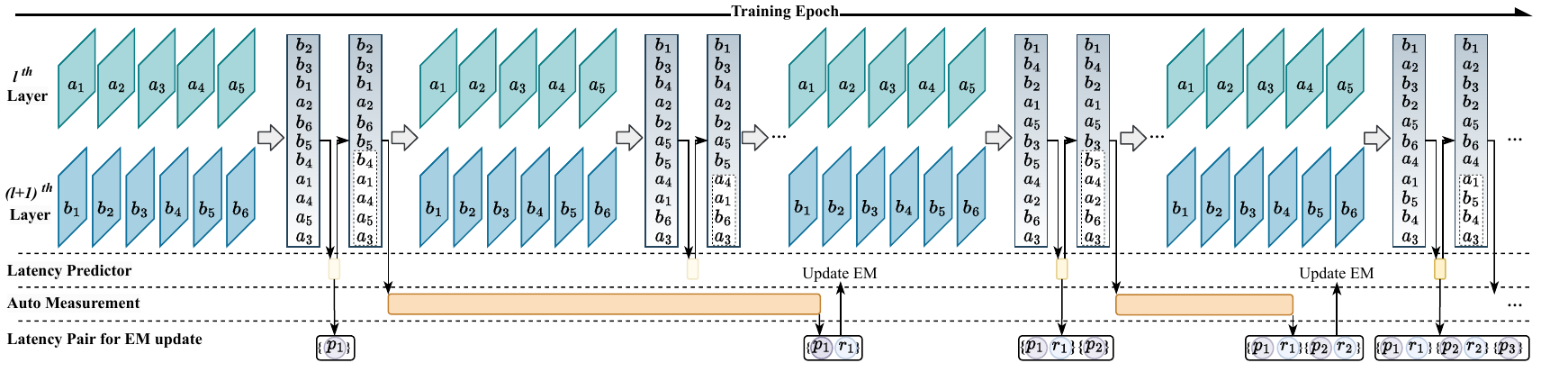}
    \caption{Latency prediction in the model compression process, including its evolving.}
    \label{fig:inpruning}
\end{figure*}

Subsequently, we generate a latency predictor for single layers by training a lightweight three-layer MLP neural network \cite{riedmiller2014multi, huai2023latency}. The forward propagation of this network is shown in Eq.~(\ref{bp}), where $x$ denotes the input vector, $W^L$ represents the weight matrix, $L$ is the index of layers, and $\sigma^L$ is the activation function at layer $L$. Due to the simplicity of this network, only a few hundred multiplication and addition operations are enough to obtain the predicted latency. The updating scheme of the MLP network is also simple, which makes the training process efficient. {\color{black} As MLP can learn the patterns in the training data \cite{riedmiller2014multi}, the trained MLP can predict the latency of unseen layers by providing their configurations. Thus, we obtain a layer-wise latency predictor (LWLP).}


\vspace{-0.3cm}
\begin{equation}
\label{bp}
   g(x) = \sigma^2(W^2\sigma^{1}(W^{1} \sigma^0(W^0x))) 
\end{equation}
\vspace{-0.5cm}
\begingroup
\color{black}
 \begin{equation}
\label{em}
\begin{aligned}
  P = \sum_{i=1}^{n} ReLU(T_i + \sum_{m=0}^{k}a_{im}*T_i^m)
    \end{aligned}
\end{equation}   
\endgroup
\vspace{-0.2cm}


To predict the overall latency of the entire model when compressed, we first get the real latency of all const layers (CL). Then, for each variable layer,  its current configuration is passed to the LWLP to obtain its predicted latency. So far, we obtain the latency of all layers under current compression, denoted as $T_n, (n=1,2,...)$. To further improve the precision of the overall model latency prediction, an evolution matrix (EM) is employed. {\color{black}The EM is all $a_{im}$ 
in Eq.~(\ref{em}), and they
are trainable parameters for the overall predicted latency ($P$) closer to the overall real latency ($R$).}  {\color{black}This approximation is similar to Taylor expansion, and it is used to emulate the edge device runtime features.
%
%
From our experiments, the approximation is accurate enough under $k=1$.
%
%
%
Meanwhile, the \textit{ReLU} activation layer is used for predicting the parallel execution of the branch layers.} 


To provide a clearer understanding of the usage and evolution of our latency predictor, Fig. \ref{fig:inpruning} depicts the model compression process that integrates our latency predictor. As previously mentioned, compression-in-one-training is more promising and is implemented in a soft-compression way \cite{huai2023latency}. Specifically, unimportant weights are updated to 0 rather than being directly removed from the model. The purpose is that these zeroized weights are not eliminated prematurely and may become important in future training processes. Fig. \ref{fig:inpruning} illustrates the compression training process in chronological order, with the oblique rectangles representing the weights, and different colors indicating different layers. The importance factors (i.e., $a_1$, $a_2$, ..., $b_5$, $b_6$) indicate the relative significance of the weights, and weights with small importance factors are ``removed" to compress the model.

The compression algorithm first sorts the importance factors and utilizes the latency predictor to identify a suitable compression ratio that meets the latency constraints, while simultaneously recording the predicted latency ($p$). The importance factors in the white dashed box correspond to the weight that needs to be ``removed". Subsequently, an automatic script \cite{kong2021edlab} is employed to measure the real latency of the compressed model on the target edge device. It is worth noting that the latency test and model training are conducted separately on the cloud and edge device, respectively, and are executed in parallel without any mutual interference. In the next training epoch, the rank of importance factors may change, requiring the latency predictor to determine a new compression ratio. However, at this stage, the edge device still tests the latency of the previous model, thus, the automatic measurement is not triggered to obtain the real latency of the current model.

Upon completion of the on-device measurement, the pair of the real latency (i.e., $r_1$) and its corresponding predicted latency (i.e., $p_1$) is used to update the EM by stochastic gradient descent. In subsequent training epochs, the updated latency predictor is invoked to find the compression ratio. However, to ensure the convergence of the EM, we still record the uncorrected prediction latency (i.e., $p_2$) in the pair. In each update of EM, all recorded latency pairs are utilized, and this process is repeated until the end of training. After training, unimportant weights are removed to obtain a high-accuracy model that satisfies the latency constraints. The evolution process of the latency predictor also offers an additional advantage. In each epoch, the predicted latency is close to the latency constraint, which enables the predictor to become more acquainted with the device execution near the constraint by updating the EM. Consequently, our latency predictor becomes more precise in predicting latency near the constraint.

Furthermore, the proposed latency predictor is not limited to model compression, but can also be used in latency-constraint NAS, etc. {\color{black} Besides, this method can be extended beyond latency prediction and utilized to build power consumption predictors, accuracy predictors, and others for various devices. We will further explore this aspect in our future research.}
\vspace{-0.5cm}
\section{Experimental Evaluation}
\subsection{Experimental Setup}
We evaluate EvoLP on four DNNs (VGG, ResNet, Inception, MobileNet) and on three edge devices (one CPU platform -- ARMv8 rev 3; and two GPU platforms -- 1: NVIDIA Pascal GPU; 2: NVIDIA Maxwell GPU). For a comprehensive comparison, we employ five widely-used baselines, FLOPs, FLOPsE (FLOPs predictor with evolution scheme), ZBNLP {\color{black} (utilizing MLP and more training data)} \cite{huai2023latency}, nn-Meter \cite{nnmeter}, and TALM \cite{li2021towards}, each of which can be used for estimating model latency in model compression. For FLOPs and FLOPsE, we use FLOPs with/without the evolution scheme to estimate latency, for ZBNLP, we use its open-source code. And we refer to the results reported in the papers of nn-Meter and TALM for comparison.  To evaluate the prediction performance, we employ three widely-used metrics in regression: the Root Mean Square Error (RMSE), the relative Root Mean Square Percentage Error (RMSPE), and the ±5\% accuracy (the proportion of models with predicted latency within a ±5\% error margin). Better prediction performance is indicated by lower RMSE, RMSPE, and higher ±5\% accuracy. Finally, we show the benefits of our predictor to model compression, in terms of model accuracy and inference latency.


\vspace{-0.5cm}
\subsection{Prediction Evaluation}
\begin{figure}[!t]
    \centering
    \includegraphics[trim={0pt 4.5pt 0pt 0pt},clip,width=0.45\textwidth]{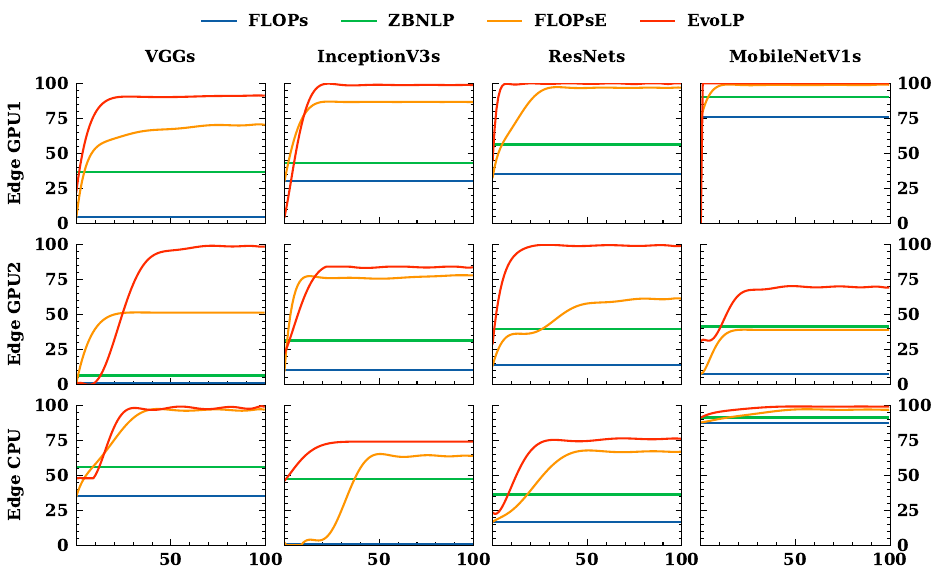}
    \caption{±5\% accuracy (Y-Axis) of different latency predictors during the training epochs (X-Axis) of the model compression process. }
    \label{fig:result1}
\end{figure}

Our proposed predictor incorporates two key techniques: the building method (i.e., utilizing MLP) and the evolution scheme during model compression. To demonstrate the distinct impacts of these two methods, we present a comparison of our approach with FLOPs, FLOPsE, and ZBNLP in a model compression process, as shown in Fig.~\ref{fig:result1}. Considering the stringent latency constraints inherent to real-time edge systems, we employ a ±5\% accuracy for evaluating these methods. From Fig.~\ref{fig:result1}, we can find that our method achieves the highest ±5\% accuracy across various devices and model variants (A variant means a model under different compression ratios, represented by \textit{+s}). {\color{black} Furthermore, the accuracy curves of ZBNLP and EvoLP reveal that the evolution scheme significantly enhances the precision of the latency predictor.  
} A comparison between FLOPsE and EvoLP demonstrates the superior effectiveness of the MLP-based method for constructing the layer predictor.


\begin{table}[!t]
\centering
\small
\caption{Latency prediction performance evaluation.}
\label{tab:t1}
   \setlength{\tabcolsep}{3pt} 
    \renewcommand{\arraystretch}{1.1} 
\begin{tabular}{lccccc}
\hline
\multicolumn{1}{c}{Method}                    & Platform                                                               & Model        & RMSE   & RMSPE & ±5\% Acc.       \\ \hline
\multicolumn{1}{c}{\multirow{8}{*}{nn-Meter}} & \multirow{4}{*}{\begin{tabular}[c]{@{}c@{}}Mobile\\ GPU\end{tabular}}  & VGGs         & 12.74  & 2.97  & 91.80           \\
\multicolumn{1}{c}{}                          &                                                                        & GoogleNets   & 0.44   & 1.35  & 100             \\
\multicolumn{1}{c}{}                          &                                                                        & ResNets      & 2.58   & 3.16  & 88.80           \\
\multicolumn{1}{c}{}                          &                                                                        & MobileNetV1s & 0.37   & 2.56  & 96.90           \\ \cline{2-6} 
\multicolumn{1}{c}{}                 & \multirow{4}{*}{\begin{tabular}[c]{@{}c@{}}Mobile\\ CPU\end{tabular}}  & VGGs         & 185.71 & 4.84  & 66.10           \\
\multicolumn{1}{c}{}                          &                                                                        & GoogleNets   & 5.69   & 3.27  & 85.90           \\
\multicolumn{1}{c}{}                          &                                                                        & ResNets      & 26.87  & 4.41  & 72.30           \\
\multicolumn{1}{c}{}                          &                                                                        & MobileNetV1s & 3.71   & 4.98  & 84.50           \\        \hline
\multicolumn{1}{c}{\multirow{4}{*}{TALM}}     & \multirow{2}{*}{\begin{tabular}[c]{@{}c@{}}Desktop\\ GPU\end{tabular}} & Inceptions   & 0.03   & -     & $\le$ 91.22 \\
\multicolumn{1}{c}{}                          &                                                                        & ResNets      & 0.10   & -     & $\le$  91.34 \\ \cline{2-6} 
\multicolumn{1}{c}{}                          & \begin{tabular}[c]{@{}c@{}}Edge\\ GPU\end{tabular}                     & Inceptions   & 6.07   & -     & $\le$  59.60 \\ \hline
\multirow{12}{*}{EvoLP}                       & \multirow{4}{*}{\begin{tabular}[c]{@{}c@{}}Edge\\ GPU 1\end{tabular}}  & VGGs         & 1.98   & 3.61  & 90.10           \\
                                              &                                                                        & Inceptions   & 1.04   & 2.27  & 99.70           \\
                                              &                                                                        & ResNets      & 0.28   & 0.93  & 99.90           \\
                                              &                                                                        & MobileNetV1s & 0.23   & 2.17  & 99.70           \\ \cline{2-6} 
                                              & \multirow{4}{*}{\begin{tabular}[c]{@{}c@{}}Edge\\ GPU 2\end{tabular}}  & VGGs         & 2.09   & 2.22  & 98.60           \\
                                              &                                                                        & Inceptions   & 2.28   & 3.69  & 83.70           \\
                                              &                                                                        & ResNets      & 1.53   & 2.34  & 99.00           \\
                                              &                                                                        & MobileNetV1s & 0.98   & 6.07  & 69.50           \\ \cline{2-6} 
                                              & \multirow{4}{*}{\begin{tabular}[c]{@{}c@{}}Edge\\ CPU\end{tabular}}    & VGGs         & 17.52  & 1.98  & 99.00           \\
                                              &                                                                        & Inceptions   & 37.42  & 10.55 & 74.00           \\
                                              &                                                                        & ResNets      & 35.73  & 7.96  & 76.00           \\
                                              &                                                                        & MobileNetV1s & 61.67  & 2.25  & 99.00           \\ \hline
\end{tabular}
\end{table} 

\begin{table}[!t]
\centering
\small

\caption{Accuracy \& Latency from Model Compression with EvoLP.}
\label{tab:t2}
   \setlength{\tabcolsep}{3pt} 
    \renewcommand{\arraystretch}{1.1} 
\begin{tabular}{cccccc}
\hline
Platform                                                              & Model        & Baseline                                                 & Comp. 1                                                 & Comp. 2                                                 & Comp. 3                                                 \\ \hline
\multirow{5}{*}{\begin{tabular}[c]{@{}c@{}}Edge\\ GPU 1\end{tabular}} & VGGs         & \begin{tabular}[c]{@{}c@{}}\textit{115 ms}\\ 84.54\%\end{tabular} & \begin{tabular}[c]{@{}c@{}}\textit{40 ms}\\ 82.64\%\end{tabular} & \begin{tabular}[c]{@{}c@{}}\textit{35 ms}\\ 81.88\%\end{tabular} & -                                                       \\ \cline{2-6} 
                                                                      & InceptionV3s & \begin{tabular}[c]{@{}c@{}}\textit{82 ms}\\ 85.62\%\end{tabular}  & \begin{tabular}[c]{@{}c@{}}\textit{60 ms}\\ 85.50\%\end{tabular} & \begin{tabular}[c]{@{}c@{}}\textit{55 ms}\\ 84.76\%\end{tabular} & \begin{tabular}[c]{@{}c@{}}\textit{50 ms}\\ 84.44\%\end{tabular} \\ \cline{2-6} 
                                                                      & ResNets      & \begin{tabular}[c]{@{}c@{}}\textit{43 ms}\\ 83.56\%\end{tabular}  & \begin{tabular}[c]{@{}c@{}}\textit{35 ms}\\ 83.88\%\end{tabular} & \begin{tabular}[c]{@{}c@{}}\textit{30 ms}\\ 83.18\%\end{tabular} & \begin{tabular}[c]{@{}c@{}}\textit{25 ms}\\ 82.24\%\end{tabular} \\ \hline
\multirow{7}{*}{\begin{tabular}[c]{@{}c@{}}Egde\\ GPU 2\end{tabular}} & VGGs         & \begin{tabular}[c]{@{}c@{}}\textit{181 ms}\\ 84.54\%\end{tabular} & \begin{tabular}[c]{@{}c@{}}\textit{40 ms}\\ 76.34\%\end{tabular} & -                                                       & -                                                       \\ \cline{2-6} 
                                                                      & InceptionV3s & \begin{tabular}[c]{@{}c@{}}\textit{125 ms}\\ 85.62\%\end{tabular} & \begin{tabular}[c]{@{}c@{}}\textit{70 ms}\\ 84.84\%\end{tabular} & \begin{tabular}[c]{@{}c@{}}\textit{60 ms}\\ 82.70\%\end{tabular} & -                                                       \\ \cline{2-6} 
                                                                      & ResNets      & \begin{tabular}[c]{@{}c@{}}\textit{88 ms}\\ 83.56\%\end{tabular}  & \begin{tabular}[c]{@{}c@{}}\textit{45 ms}\\ 82.54\%\end{tabular} & \begin{tabular}[c]{@{}c@{}}\textit{40 ms}\\ 81.62\%\end{tabular} & -                                                       \\ \cline{2-6} 
                                                                      & MobileNetV1s & \begin{tabular}[c]{@{}c@{}}\textit{23 ms}\\ 82.62\%\end{tabular}  & \begin{tabular}[c]{@{}c@{}}\textit{20 ms}\\ 81.96\%\end{tabular} & -                                                       & -                                                       \\ \hline
\end{tabular}
\normalsize
\end{table}

In addition, we present an evaluation of the final EvoLP, compared with nn-Meter and TALM (TALM only reports ±10\% accuracy) in TABLE \ref{tab:t1}\footnote{\color{black} All methods are designed for latency prediction on embedded devices.}. The results from TALM show that predicting latency for Edge GPUs is more challenging compared to Desktop GPUs. Compared to the results of TALM for Edge GPUs, we can find that our approach attains considerably higher ±5\% accuracy and lower RMSE values. Meanwhile, when compared to nn-Meter, our method achieves higher ±5\% accuracy across the majority of devices and models. This further substantiates the efficacy of our approach in model compression under strict latency constraints. 

\subsection{Model Compression with EvoLP}
Finally, we integrate our latency predictor into a model compression framework and present the final model accuracy under varying latency constraints on diverse devices using the ImageNet-100 dataset \cite{huai2023latency}, as shown in TABLE \ref{tab:t2}. The upper part in this table represents the ``hard" latency constraint, and the lower part is the achieved model accuracy while meeting the constraint by model compression with our latency predictor. The term ``Baseline" represents the model without compression, while ``Comp." is the model compressed under different latency constraints. Due to the latency predictor's error, we employ real latency measurements in the final epochs of the model compression process to ensure meeting latency constraints. TABLE \ref{tab:t2} shows that, for a specific model, larger latency corresponds to higher accuracy, suggesting that it is crucial to approach the latency constraint as closely as possible. Consequently, our precise latency predictor enables the model compression framework to be close to latency constraints, achieving higher model accuracy in one training.

\vspace{-0.3cm}
\section{Conclusion}
In this letter, we proposed a novel framework called EvoLP, which utilized an MLP neural network and self-evolution scheme to accurately predict the model inference latency on targeted hardware. EvoLP can automatically analyze the model to generate a small number of samples for measuring, thus accelerating the building process and enhancing the prediction precision. By introducing the EM, EvoLP can evolve during the compression process to further improve precision, particularly near the latency constraint. Evaluation results showed that EvoLP achieved much higher prediction precision compared to existing approaches. We also demonstrated the benefits and effectiveness of EvoLP in the model compression technique. EvoLP accurately predicted the on-device latency, facilitating the deployment of DNNs on real-time edge systems.

\bibliographystyle{IEEEtran}
\bibliography{reference}
\end{document}